\documentclass[letterpaper, 10 pt, conference]{ieeeconf}

\IEEEoverridecommandlockouts                            

\usepackage[english]{babel}
\usepackage{amsmath} 
\usepackage{amssymb}  
\usepackage{graphicx}
\usepackage{subfigure}
\usepackage{caption}
\usepackage{multirow}
\usepackage{amsfonts}
\usepackage{hyperref}
\usepackage{tabularx}
\usepackage[noend]{algpseudocode}
\usepackage{algorithm}
\usepackage{bm}
\usepackage{breakurl}
\usepackage{enumerate}
\usepackage{verbatim}
\usepackage{float}
\usepackage{siunitx}
\usepackage{mdframed}
\usepackage{adjustbox}
\usepackage{cleveref}
\usepackage{authblk}
\usepackage{color}

\usepackage{booktabs}  

\usepackage{makecell}
\usepackage[most]{tcolorbox}
\newtcolorbox[auto counter]{key}[1][]{%
    enhanced,
    breakable,
    colback=white,
    colbacktitle=white,
    coltitle=black,
    fonttitle=\bfseries,
    boxrule=.2pt,
    titlerule=.05pt,
    toptitle=2pt,
    title=Key Terminology,
    }

\title{\LARGE \bf
3D Skeletonization of Complex Grapevines for Robotic Pruning
}

\author{
Eric Schneider, Sushanth Jayanth, Abhisesh Silwal, George Kantor
\thanks{All authors are with the Carnegie Mellon Robotics Institute, PA, USA 
        \texttt{\{franzs, sushantj, asilwal, kantor\}@cs.cmu.edu}}
}

\begin{document}
\maketitle
\thispagestyle{empty}
\pagestyle{empty}


\begin{abstract}

    Robotic pruning of dormant grapevines is an area of active research in order to promote vine balance and grape quality, but so far robotic efforts have largely focused on planar, simplified vines not representative of commercial vineyards. This paper aims to advance the robotic perception capabilities necessary for pruning in denser and more complex vine structures by extending plant skeletonization techniques.
    The proposed pipeline generates skeletal grapevine models that have lower reprojection error and higher connectivity than baseline algorithms. We also show how 3D and skeletal information enables prediction accuracy of pruning weight for dense vines surpassing prior work, where pruning weight is an important vine metric influencing pruning site selection.

\end{abstract}


\section{Introduction}

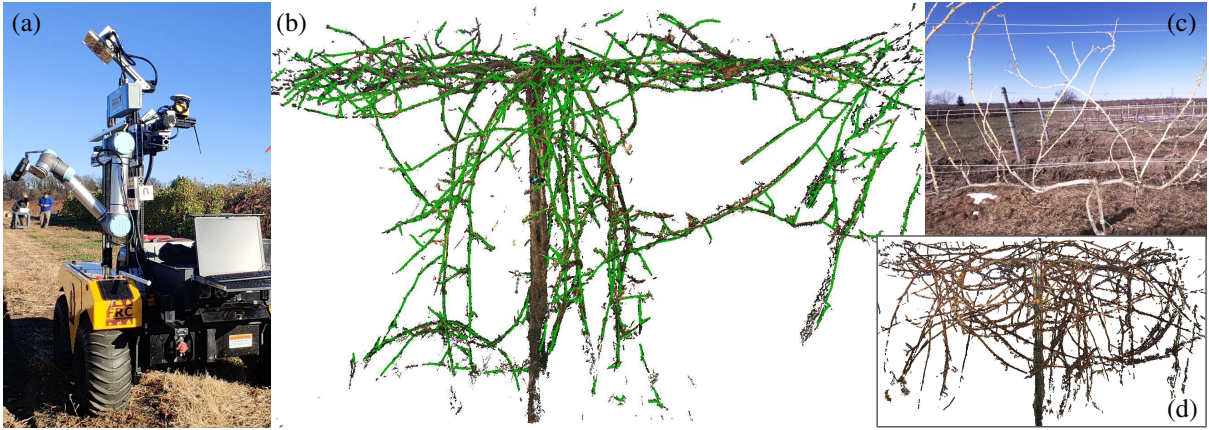
\begin{figure*}
    \centering
    \includegraphics[width=0.9\linewidth]{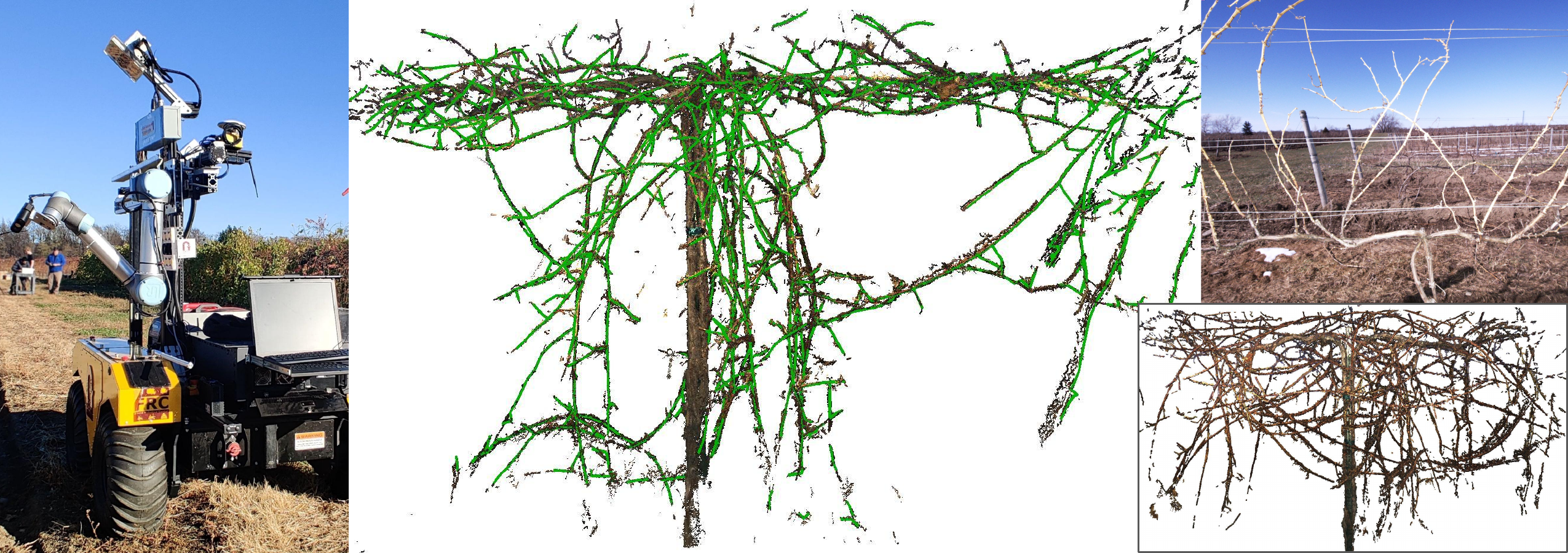}
    \put(-452, 150){\textcolor{black}{(a)}}
    \put(-352, 150){\textcolor{black}{(b)}}
    \put(-15, 150){\textcolor{black}{(c)}}
    \put(-15, 4){\textcolor{black}{(d)}}
    \caption{(a): Data collection and pruning vehicle \cite{silwal2021bumblebee}. (b) Point cloud data and overlaid skeleton (green) for a typical vine in this dataset. (c) Typical simple vine used in robotic grapevine research. Note that vines are trained in wires to roughly grow in a plane and vertically. (d) Vigorous example of vines in this dataset, scene size is \{W:3.3m, H:1.8m, D:1.2m\}.}
    \label{fig:hero}
\end{figure*}

    Vine pruning during the dormant season is an important annual operation for grape growers. It is a costly and labor-intensive process, one that growers may struggle to staff due to skilled labor shortages in agriculture. In some areas, mechanized systems have taken over as the most cost-effective solution, but lack the ability to selectively prune vines in a balanced manner.
    Robotic pruning has the potential to achieve superior outcomes compared to mechanized approaches by handling each vine according to its needs, and has been an active area of research for multiple groups \cite{botterill2017fullsystem, silwal2021bumblebee, fernandes2021grapeprune}. However, robotic pruning efforts have focused on relatively simple vines that are mostly planar, with vertically aligned growth (example in Fig.~\ref{fig:hero}(c)).

    In order to prune effectively with robots, a 3D model of the vine is required for identifying pruning locations and planning the motion of the cutting tool.
    Dense and sprawling growth, as shown in Fig.~\ref{fig:hero}(b) and (d), is difficult to model accurately. Approaches that work on sparse vines can fail to generalize due to occlusion and intertwined growth.
    We improve current plant skeletonization methods in order to produce accurate skeletons of heavy growth vines.
    In particular, we add the ability to model cycles in the skeletal model to better capture dense overlapping growth.

    In addition to producing vine skeletons, we also predict pruning weight, a measure of each vine's health and vigor. Knowing the pruning weight of vines is an important step in balance pruning \cite{smart1991sunlight}, as it is used to determine how much growth to keep or remove in the pruning process.
    We use 3D and skeletal data to predict pruning weight on dense and occluded vines more accurately than previous works.
    
    The specific contributions of this paper are:
    \begin{itemize}
        \item A skeletonization pipeline for complex vines that produces models with high reprojection scores while connecting skeletal segments
        \item A modification to graph-and-refine skeletonization strategies that handles cycles in the structure graph, allowing more accurate skeletal models
        \item Improvements on vigor estimation in dense vine growth using 3D data and skeleton information
    \end{itemize}

    We provide the images and annotations as a public dataset.

\section{Related Work}


    A common general approach for skeletonization of plants from point clouds is the graph-and-refine strategy. In \cite{xu2007treeparts, livny2010smoothertree, du2019adtree, chaudhury2020stochasticskel} the first step is to turn points into a dense graph, choose an initial single path, and then refine that path in a variety of methods. Finding a Minimum Spanning Tree (MST) path through the dense graph is a common way to start, which we adopt. At the heart of graph-and-refine processes are the ideas that physically proximate points represent connected paths in the final skeleton, and the true skeleton will be well represented by a tree graph.
    However, when processing overlapping vines, connecting nearby points causes loops that are poorly represented by a tree graph.
    In addition, it is common in graph-and-refine methods to make allometric assumptions, where the radius shrinks from the trunk through the branches in a known fashion, which does not hold for grapevines. We choose AdTree \cite{du2019adtree} as a baseline as it has open-source code to compare against.


    Skeletons are often generated for plants because cylindrical segments capture most plant growth, but skeletonization is also studied in other contexts. Laplacian Contraction (LC) \cite{cao2010cloudlaplace} is a method of general skeletonization, based on point contraction.
    By design LC on points returns a cloud without connections, where points have been compressed to the predicted skeletal axis, and does not calculate skeletal radii.

    A variety of works deal with robotic pruning.
    \cite{botterill2017fullsystem}, \cite{silwal2021bumblebee} and \cite{fernandes2021grapeprune} represent fully integrated pruning efforts that build vine models using 2D image edge tracing, region growing, and proximity-based 2D node connections, respectively. These works are evaluated on relatively simple vines, and the modelling approaches do not generalize when growth gets dense. In this paper we push perception capabilities that could allow integrated approaches to work with denser vines.

    Prior research into automated pruning weight prediction exists. In \cite{millan2019pruneweight} pruning weight is estimated from 2D cane segmentation using a monocular camera and active lighting at night.
    In \cite{kicherer2017pwestimate}, pruning weight is estimated from foreground segmentation using depth data. In both cases the vines being assessed are relatively simple and planar, and the methods do not transfer well to the dense vines in this dataset.

\section{Problem Formulation}
    In this paper, our main objective is to generate high-quality skeletons of vines, consisting of skeletal line segments in 3D space along with associated radii. Some endpoints are shared between multiple lines, indicating connectivity of the skeletal segments as seen in Fig.~\ref{fig:topology}(c). The baselines and the method presented in this paper are designed to take in a point cloud that has been semantically cropped to retain only vine points, and from that generate a skeletal model.

    Some key terminology: a \textit{vine} is one full plant, including the cordon and all individual canes. The \textit{cordon} is the oldest part of a vine, similar to a tree trunk, and a \textit{cane} is one branch of the vine, growing from the cordon and potentially splitting into further canes. \textit{Pruning weight} is the mass of all canes less than one year old cut off a single vine.







    \subsection{Dataset}
        The primary data captured consists of stereo images from side and down-facing camera pairs along a linear slider.
        Data capture was done using the robot platform from~\cite{silwal2021bumblebee}, using the flash camera system from~\cite{flashcam2021} which collects consistent images in varied outdoor lighting conditions.
        In total 144 scans were taken of Concord vines, along with pruning weight. A single scan consists of images from the two stereo pairs captured at seven positions along the linear slider. Concord vines were chosen as the most complex vines at the test site, these methods have not yet been tested on other varieties.
        Due to the high effort of annotating segmentation
        with many thin features, 91 images were labeled pixel-wise using polygons, broken into the classes (background, cane, cordon, post, leaf, sign). The images, class annotations, and pruning statistics
        are available as a dataset\footnote{Stereo Data for 144 Winter Grapevines at \url{https://labs.ri.cmu.edu/aiira/resources/}}.

\begin{figure*}
  \centering
  \includegraphics[width=0.9\linewidth]{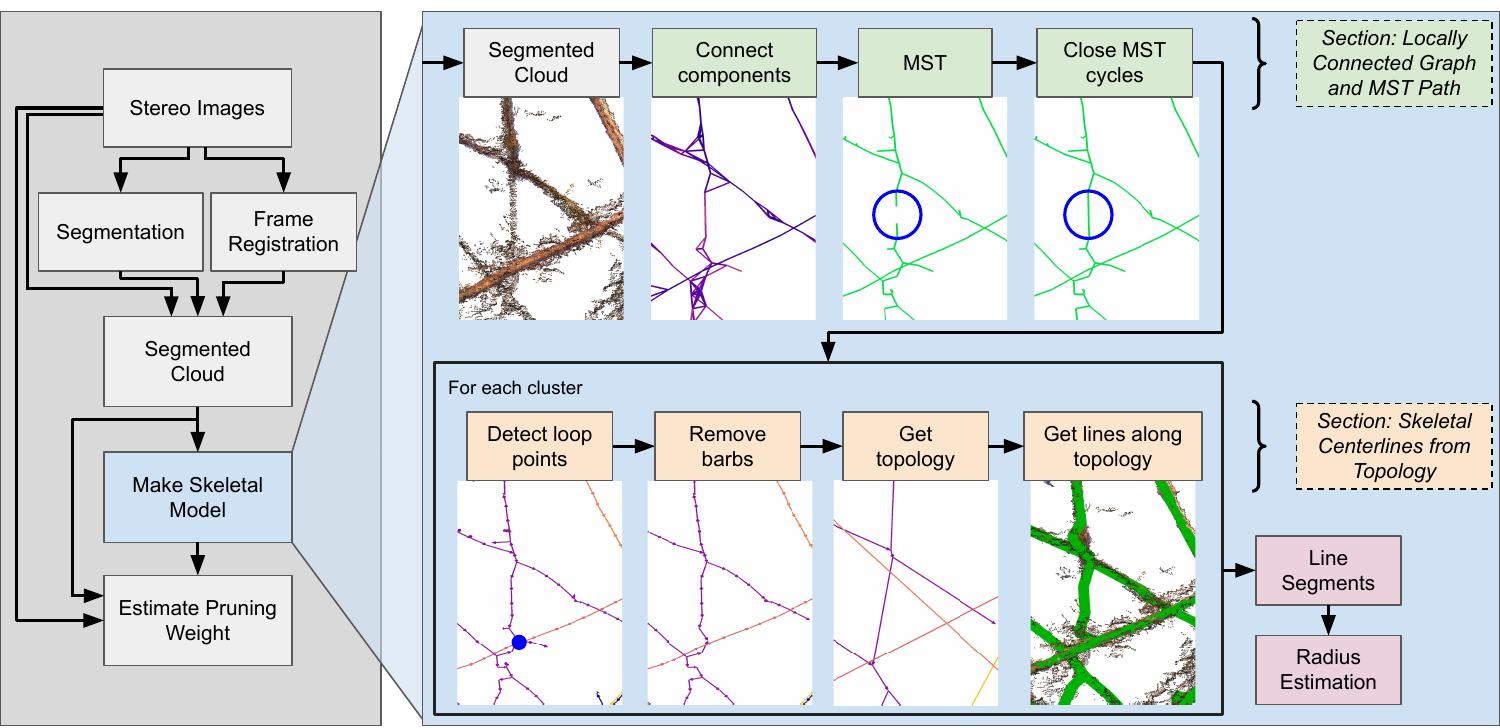}
  \caption{System diagram, showing the whole pipeline from stereo images to skeletonization and finally pruning weight estimation. Skeletal models can then be used for path planning for pruning operations.}
  \label{fig:system_diagram}
\end{figure*}

\section{Methods}

    We present a pipeline, shown in Fig.~\ref{fig:system_diagram}, which begins by taking a set of stereo images of a vine, registering the images to create a unified point cloud, and generating image segmentation masks. Then the vine-only cloud is used to build a skeletal model before combining 3D and skeletal data to predict the pruning weight for a given vine.

    \subsection{Frame to Frame Point Cloud Registration} \label{sec:register}

        After turning stereo pairs into point clouds
        and placing the cameras initially using robot extrinsics, we fine-tune camera positions similar to \cite{silwal2021bumblebee} by running Iterative Closest Point (ICP) between frames horizontally and then vertically to get a combined cloud.
        Since stereo depth error goes up quadratically with distance,
        we check regions seen by multiple cameras and discard points that come from significantly more distant cameras.



    \subsection{Cane Segmentation in Images}

        In order to separate cane points from other classes in the point cloud, we use learning-based 2D segmentation to classify pixels in the stereo images, then apply class masks onto the stereo disparity.
        We assessed various image-based segmentation models and picked the most performant.

        \subsubsection{Model and Augmentations}

            Using the MMLab segmentation toolkit \cite{mmseg2020}, we tested a series of models.
            We also tested geometric augmentations (random blur, rotation, resizing, perspective warping) and photometric augmentations (random brightness, contrast, saturation, and hue changes). The models tried were:

            \begin{itemize}
                \item FCN \cite{fcn2014}: fully connected autoencoder-like model.
                \item UNet \cite{unet2015}: designed for simplicity and low amounts of training data.
                \item BiSeNet \cite{yu2018bisenet}: split network designed to pass semantic and spatial information along separate paths.
                \item Segformer \cite{xie2021segformer}: transformer-based architecture.
            \end{itemize}

\begin{figure}[!ht]
              \vspace{-3mm}
              \centering
              \includegraphics[width=0.85\linewidth]{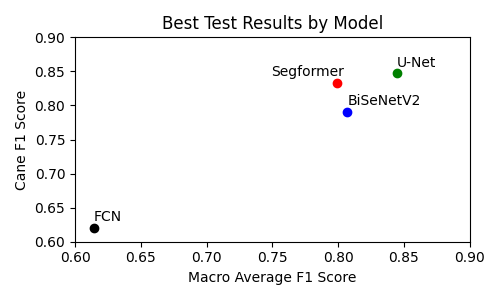}
              \caption{Model results,
              assessed with F1 score (combination of precision/recall). We use macro averaging to assess performance on all classes,
              defined as the per-class F1 mean.} 
              \label{fig:segmodels}
              \vspace{-1mm}
            \end{figure}

            The 91 labeled images were split randomly (70/20/10) into train/validate/test sets. 
            The models were trained from scratch\footnote{Except for Segformer, which comes pre-trained on ImageNet-1K} for 125 epochs.
            As seen in Fig.~\ref{fig:segmodels}, UNet has the highest performance among the tested models.
            We found that geometric augmentations improved performance, while photometric augmentations decreased it. This is likely because the flash images have such a tight domain that photometric augmentation pushed the training images outside of the domain seen in validation/test images. For segmentation we therefore use UNet with geometric augmentations.

        \subsubsection{Dilation for Precision} \label{sec:skeldilate}

\begin{figure}[!ht]
              \vspace{-2mm}
              \centering
              \includegraphics[width=\linewidth]{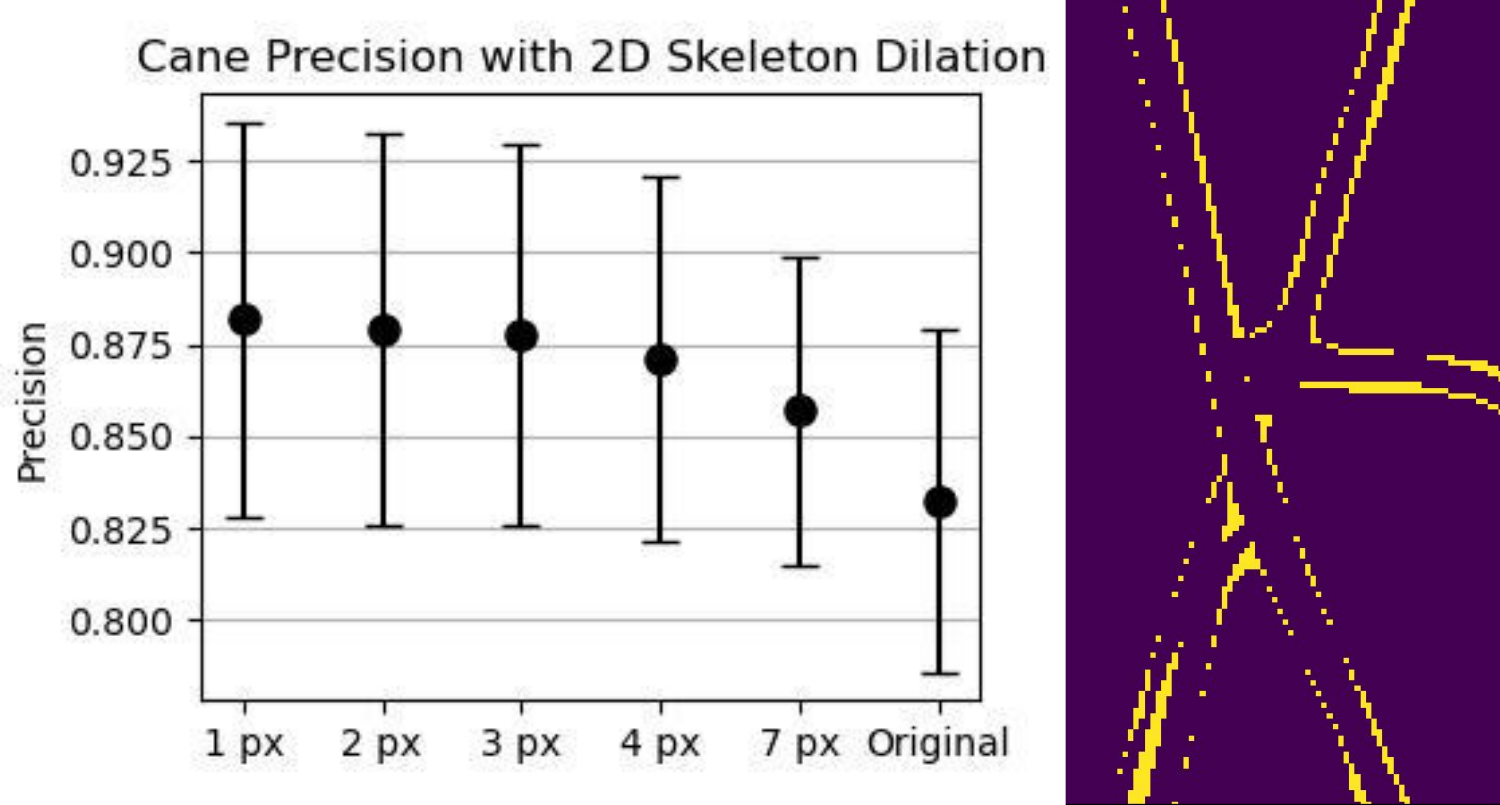}
              \put(-245, 119){\textcolor{black}{(a)}}
              \put(-13, 121){\textcolor{white}{(b)}}
              \caption{(a) Improving cane precision by discarding edge predictions, showing the mean and std. dev. From the original to the 3px cutoff the precision improved from 0.832 to 0.878, +4.6\%. (b) Segmentation error example (yellow indicates classification error), most errors are at the cane edge.} 
              \label{fig:caneprecision}
              \vspace{-4mm}
            \end{figure}

            Using UNet, we noticed a pattern in cane segmentation errors where the centers were largely correct, with errors at the cane edges.
            For skeletonization we only care about tracing cane centers, so we improved performance in the regions of interest by discarding cane edges. This is done by getting the 2D skeleton of the segmentation,
            growing that 2D skeleton 
            with dilation, then taking the intersection of the dilated shape with the original so we do not expand outside the original mask.
            Precision improvements were assessed on test set images in Fig.~\ref{fig:caneprecision}.

    \subsection{Make Skeletal Model}

        Our skeletonization approach draws on other graph-and-refine methods, and is essentially a two-part process. First a dense starter graph is created and a starting path is found through each connected cluster (\ref{sec:startergraph}), then pathways are traced and turned into line segments (\ref{sec:topo}). An overview of these steps with visualizations is shown in Fig.~\ref{fig:system_diagram}.

        \subsubsection{Locally Connected Graph and MST Path} \label{sec:startergraph}

            Given a segmented point cloud, a locally connected graph and initial MST path create a starting point for skeleton generation.



            \textbf{Connect components:} We consider all points sufficiently close as candidates for skeletal connectivity. This is accomplished by sweeping a sphere of radius $r_s$ across each point in a downsampled cloud, and building a graph where all points within the radius of point $p_i$ are connected to $p_i$.
            Neighbor querying is accomplished efficiently using a k-d tree.


            \textbf{MST:} For each connected cluster in the locally connected graph, we find an MST using the Kruskal algorithm,
            where Euclidean distance is the edge cost.

            \textbf{Close MST cycles:} By its construction, the locally connected graph will have edges wherever canes pass closely, which leads to many graph loops when canes drape over each other.
            The MST, by its nature as a tree graph, breaks these loops while minimizing path length. However, we found that broken loops led to poor skeletons because the broken loop halves would either get pruned (\ref{sec:topo}: Remove barbs) or fit as separate branches with a disconnect.
            We close the loops broken by the MST by finding leaves of the MST graph where a single step in the locally connected graph connects to another leaf, then adding that edge back.
            In order to prevent nearby barbs from connecting in tiny loops, a pre-closure graph distance of $\delta_l$ is required between the two leaves.

        \subsubsection{Skeletal Centerlines from Topology} \label{sec:topo}

            Now that we have an initial path, a series of steps are performed to generate skeletal line segments. Because the starter graph is formed by sweeping a sphere of size $r_s$ across the cloud, any gap larger than $r_s$ will cause separated clusters to form. The following steps are done independently on each cluster.

\begin{figure}[!ht]
              \centering
              \includegraphics[width=0.85\linewidth]{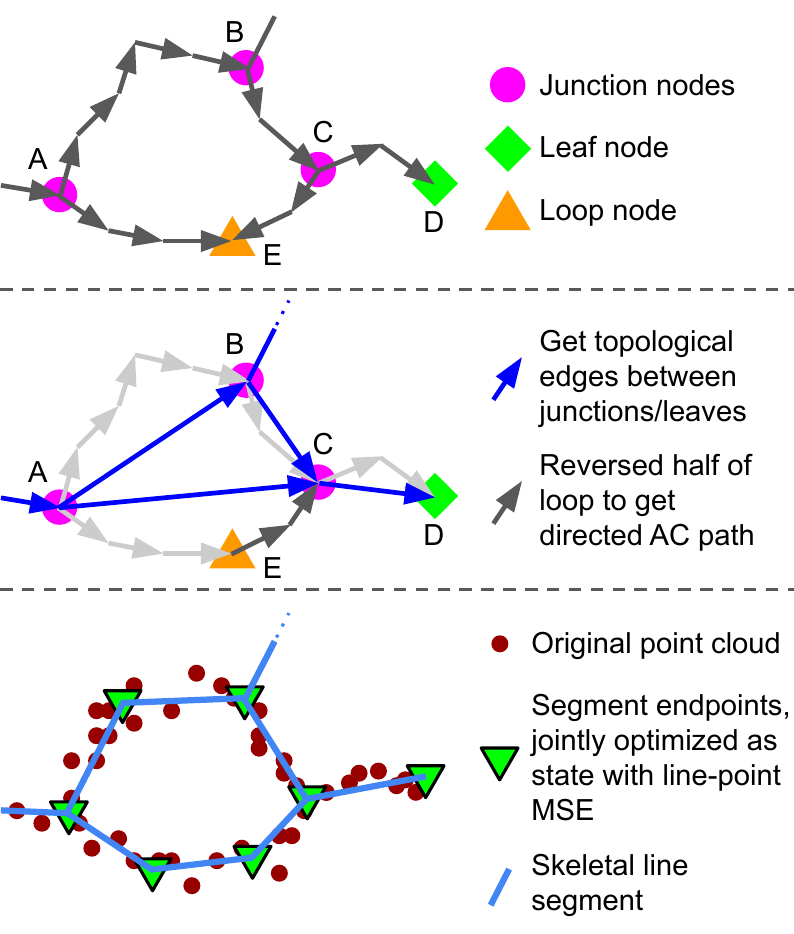}
              \put(-205, 232){\textcolor{black}{(a)}}
              \put(-205, 156){\textcolor{black}{(b)}}
              \put(-205, 78){\textcolor{black}{(c)}}
              \caption{(a) Smooth graph from \textbf{Remove barbs}, with junctions and a leaf. (b) In \textbf{Get topology}, mid-nodes are eliminated. The shorter loop half, CE, is reversed to get a continuous AC path. (c) In \textbf{Get lines along topology}, when fitting lines to the points, the state is the positions of all green triangles.} 
              \label{fig:topology}
              \vspace{-2mm}
            \end{figure}

            \textbf{Detect loop points:}
            In order to ensure the directed graph maintains loops, loop points are detected. Any nodes that are the common endpoint of two or more directed edges are saved as a loop point, as demonstrated in Fig.~\ref{fig:topology}(a).

            \textbf{Remove barbs:} MSTs of the locally connected graph form long paths with small offshoots,
            much like barbed wire. Inspired by \cite{du2019adtree}, we remove small barbs by removing nodes where the downstream edge length is
            below a threshold $\delta_b$.

            \textbf{Get topology:} from the smooth graph we identify isolated cane sections by eliminating all but junction/leaf nodes as illustrated in Fig.~\ref{fig:topology}(b). All mid-nodes with only one parent and one child are discarded, leaving a topological graph where each edge represents a single cane of variable length. In order to handle loops, loop points with two incoming edges are treated as mid-nodes to eliminate, where the shorter side of the loop is reversed so that when the loop is collapsed the topological edge reaches from one junction to the other.

            \textbf{Get lines along topology:} given
            topological edges representing a non-branching stretch of cane, skeletal line segments
            are generated. This is done by associating the original points
            to a single topological edge, then fitting line segments to minimize the point-to-line Mean Squared Error (MSE). The $(x, y, z)$ values of all line endpoints form the state when concatenated, so batches of line ends are optimized jointly as shown in Fig.~\ref{fig:topology}(c). It is important to optimize over
            shared endpoints to preserve connectivity.
            When the connected cluster is too large,
            this process is performed on segments of the cluster.

        \subsubsection{Radius Estimation}
    
            Finally, after finding skeletal center lines, the radii of all canes in a cluster ($r_i \in R$) are estimated jointly using linear regression.
            Three aspects are balanced to determine the radii: a prior value, a smoothing term, and point-fitting. For the prior, we set the radius for a given line $r_i$ equal to the prior radius, $r_\text{prior}$. For smoothing, for every pair of line segments $(l_i, l_j)$ that share a junction the radii are set equal: $r_i = r_j$. Finally, each point $p_k$ associated with a given line segment $l_i$ is set so that the distance $\delta_k$ from $p_k$ to $l_i$ is equal to $r_i$.
            Here is the linear system in matrix form:

            \begin{figure}[!ht]
              \centering
              \includegraphics[width=0.85\linewidth]{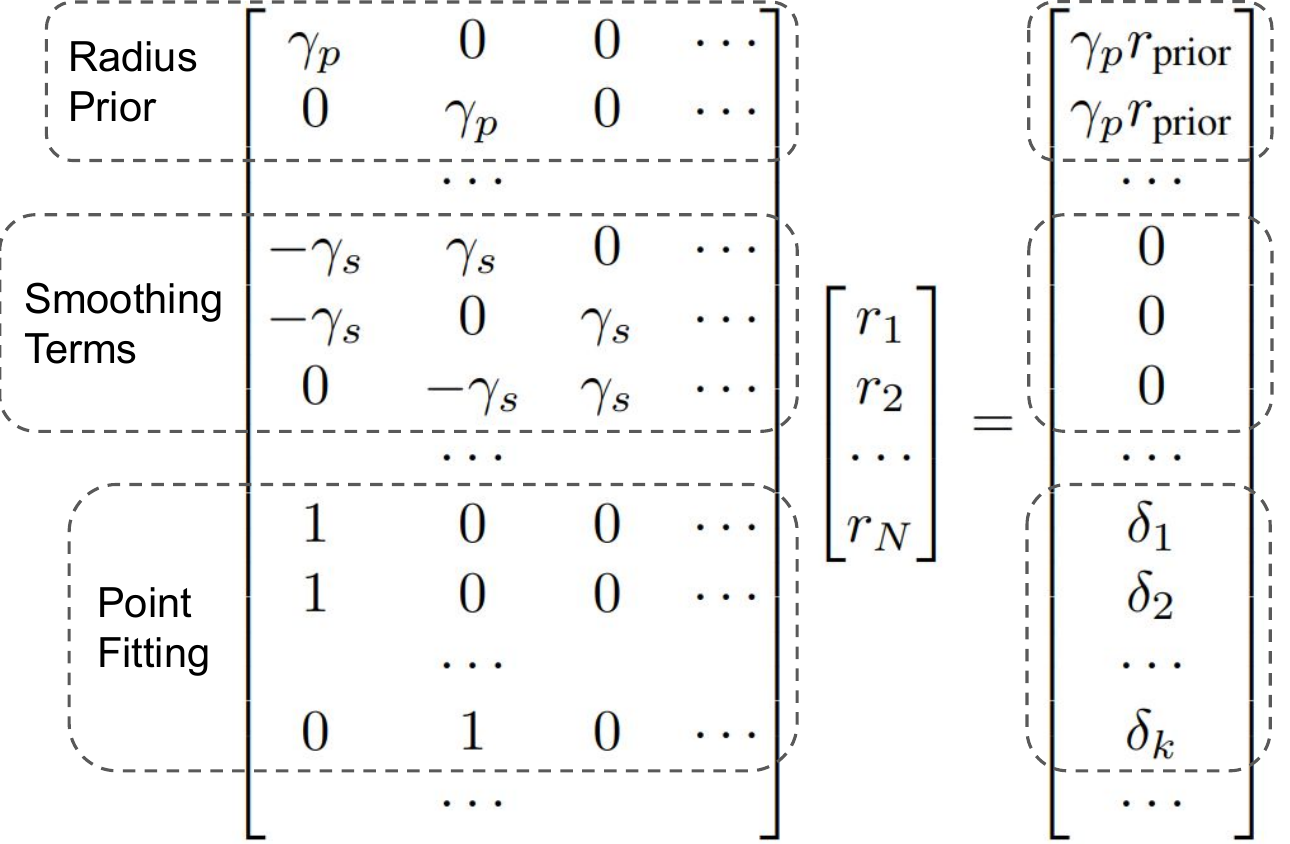}
              \vspace{-3mm}
            \end{figure}

            $\gamma_p$ and $\gamma_s$ are weights for the prior value and smoothing terms, scaled by $k = \frac{|P|}{|R|}$, the average points per radius. The best results were with $\gamma_p=k, \gamma_s=0.1k$, with $r_\text{prior} = 5$mm.


    \subsection{Pruning Weight Estimation}

        After skeletonization, we calculate pruning weight using linear regression on five variables available from each vine.
        Z-score normalization is used so variable magnitudes are balanced. The variables are:

        \begin{itemize}
            \item Cane voxels: number of filled voxels after voxel downsampling the cane cloud at a voxel size of 2cm.
            \item Cordon voxels: number of filled voxels after voxel downsampling the cordon cloud at a voxel size of 2cm.
            \item Pole distance: average distance from the robot to the central pole the vine grows on.
            \item Skeleton length: sum of line segments in the skeleton.
            \item Cane pixels: total number of pixels segmented as cane.
        \end{itemize}

        We chose a linear model because of the low number of data points. With more data a 2D or 3D learning model with higher capacity
        could produce better results, but for small datasets a low capacity model is simple to implement and prevents overfitting.
\section{Experimental Results}~\label{sec:experiments}

    \vspace{-3mm}
    \subsection{Skeletonization Results} \label{sec:skelresults}

\begin{figure}[!ht]
          \vspace{-1mm}
          \centering
          \includegraphics[width=\linewidth]{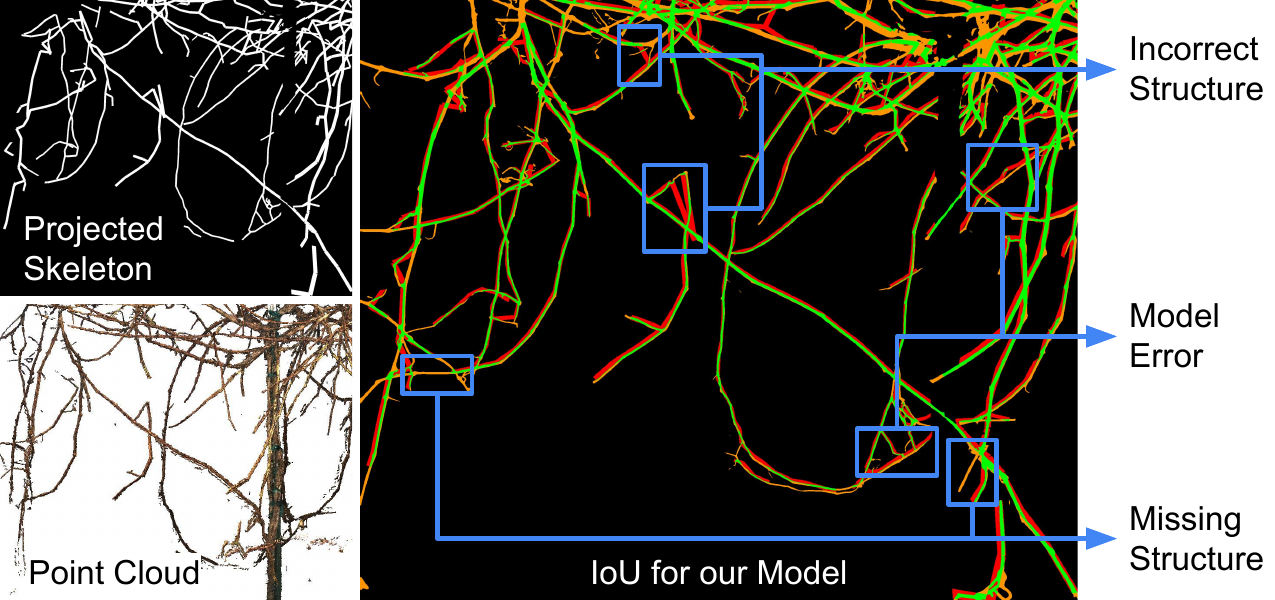}
          \caption{Skeleton quality is assessed by projecting the skeleton
          onto cane segmentation. \textbf{Green}: match between model, segmentation. \textbf{Orange}: cane segmentation with no projected model. \textbf{Red}: projected model with no segmentation.} 
          \label{fig:iou_explainer}
          \vspace{-2mm}
        \end{figure}

        We adopt the unsupervised skeletal reconstruction metric from \cite{botterill2017fullsystem}, which uses Intersection over Union (IoU) of the model projected onto a segmentation mask.
        As shown in Fig.~\ref{fig:iou_explainer}, IoU checks whether the skeletal model covers areas classified as cane.
        We also assess the number of connected clusters. In general more clusters means a more fragmented skeleton, which provides less connectivity information.

\begin{figure}[!ht]
          \centering
          \includegraphics[width=\linewidth]{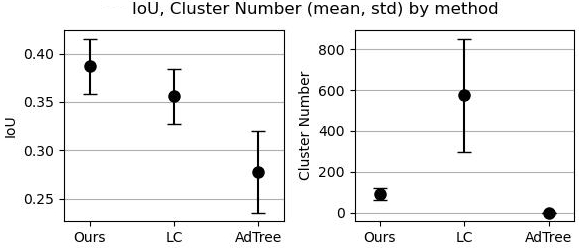}
          \caption{Reprojection scores and cluster count across methods.}
          \label{fig:ours_v_base}
          \vspace{-2mm}
        \end{figure}

        Fig.~\ref{fig:ours_v_base} shows the results of our offline skeletonization method against two offline baselines with open source code, Laplacian Contraction \cite{cao2010cloudlaplace} and AdTree \cite{du2019adtree}. We produce skeletons that recreate the visible vine structure more accurately, while providing higher vine connectivity than Laplacian Contraction. Although AdTree is more connected than our approach, in this context AdTree's assumption that all points connect is too strong, leading to forced connections to the central cordon that do not exist.


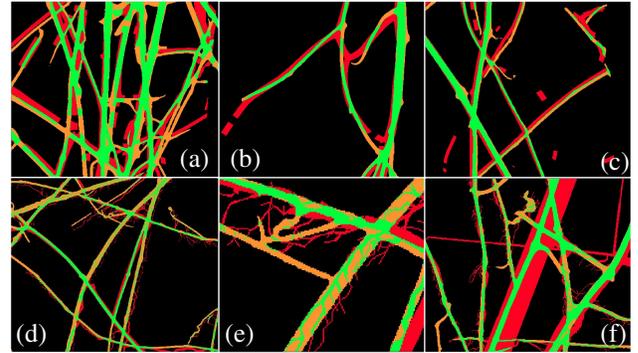
\begin{figure}[!ht]
          \vspace{-2mm}
          \centering
          \includegraphics[width=0.96\linewidth]{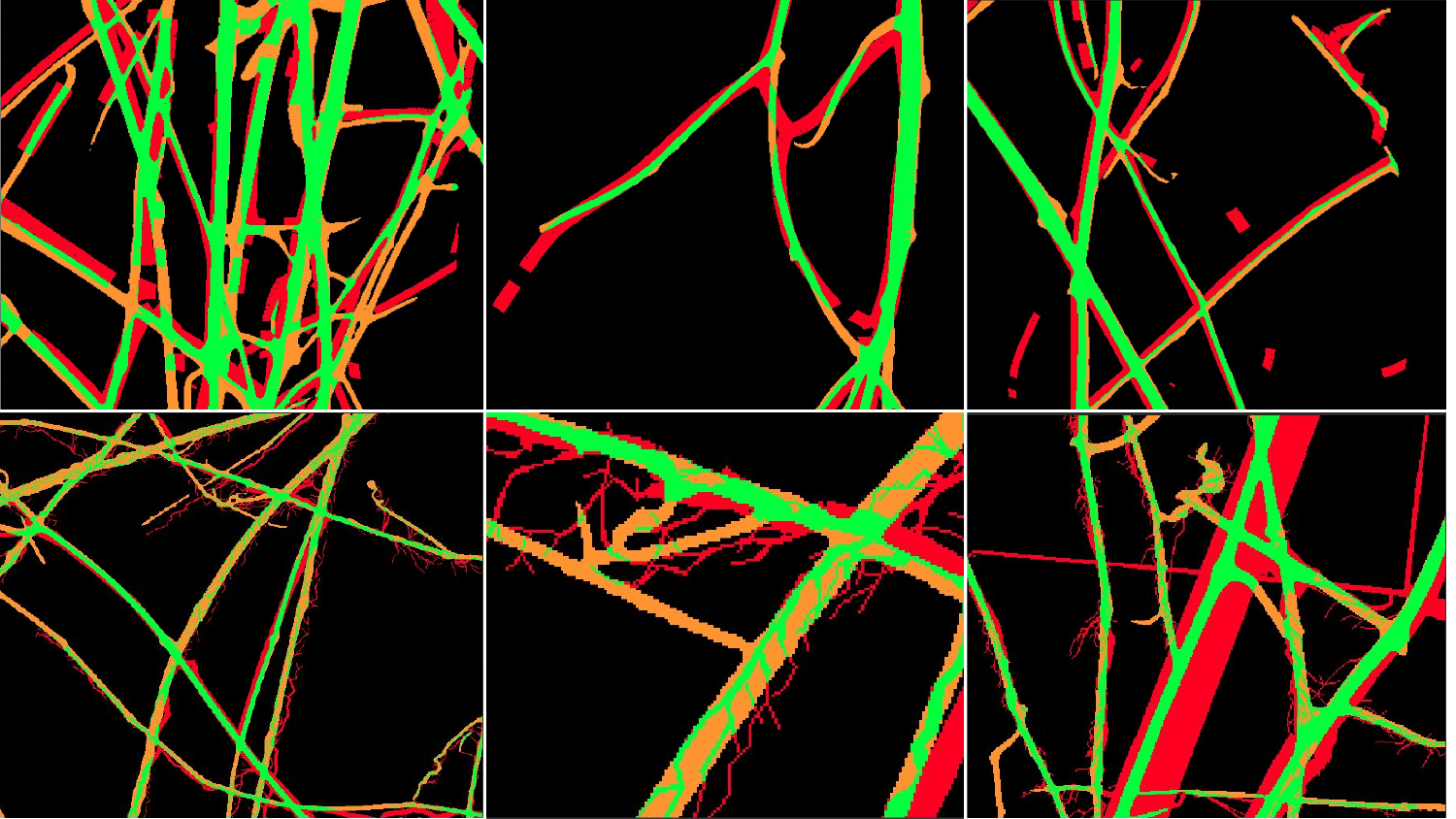}
          \put(-172, 70){\textcolor{white}{(a)}}
          \put(-153, 70){\textcolor{white}{(b)}}
          \put(-13, 69){\textcolor{white}{(c)}}
          \put(-234, 4){\textcolor{white}{(d)}}
          \put(-155, 4){\textcolor{white}{(e)}}
          \put(-13, 4){\textcolor{white}{(f)}}
          \caption{Qualitative plots,
          colors explained in Fig.~\ref{fig:iou_explainer}. In open spaces Laplacian Contraction (a-c) is drawn correctly onto the cane structure, but junctions pull neighboring points awry (b), and it leads to fragmentation (c). AdTree (d-f) has two primary issues,
          filaments (e) and over-estimating radius (f) due to allometric assumptions.}
          \label{fig:qualitative_skels}
          \vspace{-2mm}
        \end{figure}

        Since Laplacian Contraction only returns contracted points, to evaluate it we assume each contracted point is connected to its two nearest neighbors and has a fixed radius.
        Skeletons formed by this method are relatively fragmented, with no method of joining likely paths, which makes them less useful for robotic pruning use cases. Qualitative views found in Fig.~\ref{fig:qualitative_skels} (a-c).


        AdTree represents the opposite extreme, where each vine model consists of a single cluster.
        In a complex vineyard setting, neighboring vines grow into the space, and assuming those canes connect to the central cordon degrades the IoU.
        In addition, AdTree uses allometric tree growth assumptions to calculate radii, and grapevines do not follow the same patterns as trees. Radii are therefore over-estimated near the cordon and unrealistic hair-like tendrils are formed at the tip.
        Qualitative views of AdTree can be seen in Fig.~\ref{fig:qualitative_skels} (d-f).

    \subsection{Pruning Weight Prediction Results}

        Table~\ref{table:pwfit} contains the results of our best linear model for predicting pruning weight, compared against two prior works. We assess model quality using the coefficient of determination $R^2$, as well as the root mean squared error (RMSE) of our predicted weight vs. ground-truth. To assess stability, we do a 100-fold assessment with a (70/30) train/test split, and report standard deviation over folds.

\begin{table}[!ht]
\vspace{-2mm}
\begin{center}
\begin{tabular}{l|ll|ll|}
\cline{2-5}
                               & \multicolumn{2}{l|}{$\textbf{R}^2$}               & \multicolumn{2}{l|}{\textbf{RMSE} (kg)}             \\ \hline
\multicolumn{1}{|l|}{\textbf{Method}}   & \multicolumn{1}{l|}{Avg.} & Std. Dev. & \multicolumn{1}{l|}{Avg.} & Std. Dev. \\ \hline
\multicolumn{1}{|l|}{Ours}     & \multicolumn{1}{l|}{0.51} & 0.10      & \multicolumn{1}{l|}{0.33} & 0.03      \\ \hline
\multicolumn{1}{|l|}{Cane pixel count \cite{millan2019pruneweight}} & \multicolumn{1}{l|}{0.33} & 0.10      & \multicolumn{1}{l|}{0.39} & 0.04      \\ \hline
\multicolumn{1}{|l|}{Cane surface area \cite{kicherer2017pwestimate}} & \multicolumn{1}{l|}{0.38} & 0.10      & \multicolumn{1}{l|}{0.38} & 0.04      \\ \hline
\end{tabular}
\caption{We show results from our method, the correlation of cane pixel count to PW as in \cite{millan2019pruneweight}, and the correlation of cane surface area to PW as in \cite{kicherer2017pwestimate}.} 
\label{table:pwfit}
\end{center}
\vspace{-5mm}
\end{table}

        Our pruning weight predictions shown in Table~\ref{table:pwfit} are more accurate than prior works on dense vines.
        One grower said they would consider a pruning weight system viable once the RMSE was less than (0.5lbs/0.23kg). Our system approaches that functional level, but still needs improvement.



\begin{figure} [!ht]
          \vspace{-2mm}
          \centering
          \includegraphics[width=\linewidth]{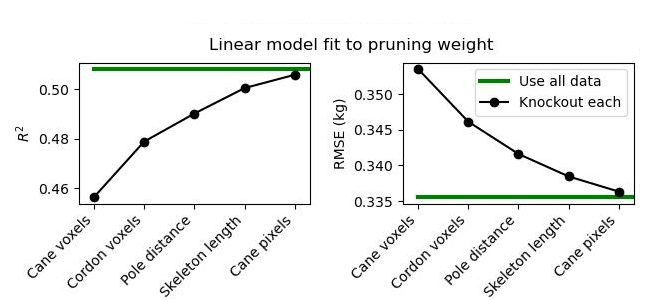}
          \caption{Effects on test data of dropping each variable from the model, also using 100 folds with a (70/30) train/test split.}
          \label{fig:pwanalysis}
          \vspace{-3mm}
        \end{figure}

        We explore the effects of dropping any one variable from the model in Fig.~\ref{fig:pwanalysis}. Our initial assumption was that skeletal length would be a strong predictor of pruning weight, however in practice it appears to be beneficial but play a smaller predictive role.


        One factor that may degrade predictive performance
        is that
        it is common for canes to grow into the spaces of neighboring vines. These neighbor canes are captured in images, but when measuring pruning weight are disentangled and assigned to the vine they originated from. Thus part of the structure detected for a given vine should theoretically be assigned to its neighbors.
        In addition, although pruning weight is defined as the mass of pruned canes less than a year old, this prediction method operates without knowledge of where canes will be cut, or cane age.
        Pruning weight prediction could be improved if a reliable method of attributing visible canes to the correct source were developed, as well as calculating accurate estimates for cut points and cane age.



\section{Conclusion}\label{sec:conclusion}

    In order to perform pruning in complex, noisy environments, it is important to understand the location and structure of canes. This work attempts to advance skeletonization efforts in dense vine structures by creating a skeletonization pipeline with a modified graph-and-refine approach, achieving higher skeletal coverage than the baselines.

    For pruning weight prediction, it would be promising for future work to use higher capacity 2D or 3D learning methods, which would necessitate larger amounts of training data. As robotic pruning gets more mature, robots could eventually predict pruning mass, prune the vine, collect pruning weight, and close the loop to improve the predictive performance in a beneficial data collection cycle.



    One fact that has become clear during this work is that while it is useful to represent the location of canes as we do with skeletal links, in reality all canes must grow from some source, and knowledge of plant growth patterns gives insight into which source a given cane comes from.
    In future work, developing a new type of plant skeletal model based on growth sources and likely growth pathways, along with methods to accurately construct those models, would be beneficial not just for pruning but for a variety of robotics challenges that deal with the dynamics and manipulability of plants, such as harvesting, grafting, and pollinating.






\section*{ACKNOWLEDGMENTS}
The authors would like to thank Terry Bates and the team at the Cornell Lake Erie Research and Extension Laboratory for allowing vine access and collecting pruning statistics, as well as the AdTree authors for baseline support. This work was partially supported by USDA NIFA/NSF National Robotics Initiative 2021-67021-35974.

\bibliographystyle{IEEEtran}
\bibliography{mybib}
\end{document}